\documentclass[runningheads]{llncs}
\usepackage[T1]{fontenc}
\usepackage{graphicx}
\usepackage{multirow}
\usepackage[utf8]{inputenc} 
\usepackage[T1]{fontenc}    
\usepackage{hyperref}       
\usepackage{url}            
\usepackage{booktabs}       
\usepackage{amsfonts}       
\usepackage{nicefrac}       
\usepackage{microtype}      
\usepackage{xcolor}         
\usepackage{times}
\usepackage{epsfig}
\usepackage{graphicx}
\usepackage{amsmath}
\usepackage{amssymb}
\usepackage{indentfirst} 
\usepackage{changepage} 
\usepackage{hyperref}
\usepackage{soul}
\usepackage{booktabs}
\usepackage{algorithm}
\usepackage{algorithmic}
\usepackage{makecell}
\usepackage{appendix}
\usepackage{subfigure}
\usepackage{authblk}
\usepackage{wrapfig}
\usepackage{siunitx}
\usepackage{booktabs}
\usepackage{makecell}

\usepackage{pifont}
\usepackage{fontawesome} 
\usepackage{bbding}
\usepackage{blindtext}
\newcommand*\samethanks[1][\value{footnote}]{\footnotemark[#1]}

\begin{document}
\title{SVFormer: A Direct Training Spiking Transformer for Efficient Video Action Recognition}

\titlerunning{SVFormer for Efficient Video Action Recognition}
%
\author{Liutao Yu\inst{1} \and
Liwei Huang\inst{1,2} \and
Chenlin Zhou\inst{1} \and
Han Zhang\inst{1,3} \and
Zhengyu Ma\inst{1}\thanks{Corresponding author} \and
Huihui Zhou\inst{1}\samethanks \and
Yonghong Tian\inst{1,2}
}

\authorrunning{L. Yu et al.}
%
\institute{AI Department, Peng Cheng Laboratory, Shenzhen, China \\ \email{\{yult, zhoucl, mazhy, zhouhh\}@pcl.ac.cn} \\ \and
National Key Laboratory for Multimedia Information Processing, School of Computer Science, Peking University, Beijing, China \\
\email{huanglw20@stu.pku.edu.cn, yhtian@pku.edu.cn}\\ \and
Faculty of Computing, Harbin Institute of Technology, Harbin, China\\
\email{23B303002@stu.hit.edu.cn}}



%
\maketitle              

\begin{abstract}
Video action recognition (VAR) plays crucial roles in various domains such as surveillance, healthcare, and industrial automation, making it highly significant for the society. Consequently, it has long been a research spot in the computer vision field.
As artificial neural networks (ANNs) are flourishing, convolution neural networks (CNNs), including 2D-CNNs and 3D-CNNs, as well as variants of the vision transformer (ViT), have shown impressive performance on VAR. However, they usually demand huge computational cost due to the large data volume and heavy information redundancy introduced by the temporal dimension. 
To address this challenge, some researchers have turned to brain-inspired spiking neural networks (SNNs), such as recurrent SNNs and ANN-converted SNNs, leveraging their inherent temporal dynamics and energy efficiency. Yet, current SNNs for VAR also encounter limitations, such as nontrivial input preprocessing, intricate network construction/training, and the need for repetitive processing of the same video clip, hindering their practical deployment. 
In this study, we innovatively propose the directly trained SVFormer (Spiking Video transFormer) for VAR. SVFormer integrates local feature extraction, global self-attention, and the intrinsic dynamics, sparsity, and spike-driven nature of SNNs, to efficiently and effectively extract spatio-temporal features. 
We evaluate SVFormer on two RGB datasets (UCF101, NTU-RGBD60) and one neuromorphic dataset (DVS128-Gesture), demonstrating comparable performance to the mainstream models in a more efficient way. Notably, SVFormer achieves a top-1 accuracy of 84.03\% with ultra-low power consumption (21 mJ/video) on UCF101, which is state-of-the-art among directly trained deep SNNs, showcasing significant advantages over prior models.

\end{abstract}

\section{Introduction}

Video is becoming a prevalent and indispensable medium to convey information in daily life, which captures movements, actions, and events over time.
Video action recognition (VAR) is an important aspect of video understanding, focusing on automatically identifying actions or activities from videos. This capability is invaluable for various applications, including surveillance, healthcare, entertainment, sports, education, industrial automation, and beyond. 
Nevertheless, video processing presents greater challenges compared to images, given the necessity to model temporal dynamics within large data volume and heavy information redundancy introduced by the temporal dimension. 

As artificial neural networks (ANNs) have shown great success in various computer vision tasks, a lot of studies based on convolutional neural networks (CNNs) \cite{chen2021deep} or vision transformers (ViTs) \cite{selva2023video} have emerged for VAR in recent years. 
In the early stages, CNNs are the mainstream approaches to VAR, including 2D-CNNs \cite{wang2016temporal,zhou2018temporal,fan2019more} and 3D-CNNs \cite{tran2015learning,carreira2017quo,hara2018can}.  
The decomposition of 3D-CNNs \cite{qiu2017learning,xie2018rethinking,tran2018closer}, as well as the combination of 2D-CNNs and 3D-CNNs \cite{feichtenhofer2019slowfast}, are adopted to reduce the computation cost. 
However, CNNs struggle to learn long-range dependency between patches, due to the inductive bias of convolution. 
To overcome this challenge, network models based on ViTs are becoming popular and showing good performance for VAR in recent years \cite{selva2023video}, including TimeSformer \cite{bertasius2021space}, ViViT \cite{arnab2021vivit}, MViT \cite{fan2021multiscale}, video swin transformer \cite{liu2022video}, and so on. To further improve the performance, methods combining convolution and self-attention \cite{liu2020convtransformer,li2022uniformer,li2023uniformer}, or using self-supervised pretraining \cite{tong2022videomae,wang2023videomae,wei2022masked} are proposed. 
Nevertheless, the huge computation cost of current ANNs for VAR still limits their practical deployment, especially in power-constrained situations.

The brain-inspired spiking neural networks (SNNs) have garnered significant attention for their potential in temporal information processing and energy efficiency \cite{roy2019towards}, thus are ideal candidates for processing videos. 
Previous studies show that SNNs exhibit good performance on various tasks, such as object recognition/detection/tracking, robotics control and so on \cite{yamazaki2022spiking,guo2023direct,Zhou2024DirectTH}. 
In recent years, recurrent SNNs (RSNNs) \cite{panda2018learning,chakraborty2023heterogeneous} and ANN-converted SNNs \cite{zhang2022high,you2024converting} have been applied to VAR. 
However, they also encounter limitations in practical application: RSNNs usually need nontrivial input preprocessing and network construction/training, as well as long simulation steps \cite{panda2018learning,chakraborty2023heterogeneous}; ANN-converted SNNs are based on well-trained ANNs, and need to process the same video clip several times to accomplish the task \cite{zhang2022high,you2024converting}. 

In this study, we innovatively propose the directly trained SVFormer (Spiking Video transFormer) for VAR, aiming to address the challenges outlined above. 
SVFormer processes a video clip frame-by-frame without the need for complex input processing, and can be trained end-to-end through the surrogate gradient method, enabling straightforward incremental learning and facilitating practical deployment.  
SVFormer integrates local feature extraction, global self-attention, and the intrinsic dynamics, sparsity, and spike-driven nature of SNNs, to efficiently and effectively extract spatio-temporal features. 
Besides, we incorporate parametric LIF neuron \cite{fang2021incorporating}, a local-global-fusion operation and a novel time-dependent batch normalization into SVFormer, contributing to its good performance. 
We first evaluate SVFormer on the classical UCF101 dataset, and achieve a top-1 accuracy of 84.03\% with ultra-low power consumption (21 mJ/video), which is state-of-the-art among directly trained deep SNNs, showing significant advantages over previous models. 
To validate its generalizability, SVFormer is then evaluated on a larger RGB datasets (NTU-RGBD60) and a neuromorphic dataset (DVS128-Gesture), both demonstrating comparable performance to the mainstream models in a more efficient way. 
These results showcase that the directly trained SVFormer is an effective and efficient model for VAR. 

\section{Related work}
\subsection{ANNs for VAR}
In the early stages, CNNs including 2D-CNNs and 3D-CNNs, are the mainstream approaches to VAR \cite{chen2021deep}. 
Some studies applied two-stream models to process RGB frames and optical flows in two separate CNNs, with a late fusion operation in deeper layers \cite{karpathy2014large,cheron2015p,christoph2016spatiotemporal,feichtenhofer2017spatiotemporal}. 
Another line of approach adopts 2D-CNNs to extract frame-level features, and then models temporal information of the input sequence in various ways, such as the consensus module in TSN \cite{wang2016temporal}, the bag of features in TRN \cite{zhou2018temporal}, the pointwise convolutions across frames in TAM \cite{fan2019more}, and so on. 
Moreover, to better model the spatio-temporal features, many variants of 3D-CNNs have been introduced, such as C3D \cite{tran2015learning}, I3D \cite{carreira2017quo} and ResNet3D \cite{hara2018can}. To mitigate the high computational burden brought by 3D convolutions, some studies tried to decompose the 3D convolution into 2D spatial convolution and 1D temporal convolution, such as P3D \cite{qiu2017learning}, S3D \cite{xie2018rethinking} and R(2+1)D \cite{tran2018closer}; or to use a combination of 2D-CNNs and 3D-CNNs, such as SlowFast \cite{feichtenhofer2019slowfast}. 
However, CNNs mainly extract local features due to limited receptive fields, ignoring long-range dependency across patches.

Recently, ViTs outperform CNNs in many visual tasks due to their enhanced ability to capture long-range dependencies \cite{dosovitskiy2020image,han2022survey}. Some researchers proposed different variants based on ViTs for VAR \cite{selva2023video}. The classical TimeSformer enables spatio-temporal feature extraction directly from a sequence of frame-level patches, and explores different space-time self-attention schemes \cite{bertasius2021space}. ViViT is a pure-transformer based model, which factorizes the spatial- and temporal-dimensions of the input to handle the long sequences of tokens \cite{arnab2021vivit}. To reduce the computation cost, Fan et al. introduced a multiscale pyramid of features and a pooling self-attention mechanism in MViT \cite{fan2021multiscale}; and Liu et al. proposed an inductive bias of locality with shifted window-based self-attention in a video swin transformer \cite{liu2022video}. Besides, the combination of convolution and self-attention is also adopted for efficiently learning video representations \cite{liu2020convtransformer,li2022uniformer,li2023uniformer}. 
Furthermore, assisted with large-scale self-supervised pretraining, great improvements have been achieved for VAR, such as in VideoMAE \cite{tong2022videomae,wang2023videomae} and MaskFeat \cite{wei2022masked}. 
Although current models have realized high classification accuracy for VAR, the excessive data volume and heavy information redundancy introduced by the temporal dimension limit their practical deployment in power-constrained situations.

\subsection{SNNs for VAR} 
Considering the intrinsic dynamics and energy efficiency of SNNs, they are ideal candidates for VAR and have been attempted in some recent work \cite{panda2018learning,chakraborty2023heterogeneous,wang2019temporal,zhang2022high,you2024converting}. 
Panda and Srinivasa developed a Driven-Autonomous based reservoir recurrent SNN to recognize video actions from limited training examples, which is not trivial to train, demonstrating a top-1 accuracy of 81.3\% on the UCF101 dataset with pre-extracted multi-scan spike sequences of 300 time steps as input to the model \cite{panda2018learning}. 
Further, Chakraborty and Mukhopadhyay presented a heterogeneous recurrent SNN (HRSNN) with unsupervised learning for VAR on several RGB datasets (KTH, 94.32\%; UCF101, 77.53\%) and one neuromorphic dataset (DVS128-Gesture, 96.54\%) \cite{chakraborty2023heterogeneous}, using a similar input preprocessing method as \cite{panda2018learning} and a nontrivial initialization method. 
Wang et al. progressively trained a two-stream hybrid network (TSRNN) consisting of CNN, RNN, and a novel spiking module, where spiking signals correct the memory of RNN, achieving competitive performance on UCF101 (94.4\%) and HMDB51 (69.9\%) with both RGB frames and optical flows as input \cite{wang2019temporal}. 
Zhang et al. constructed a two-stream deep recurrent SNN model through a hybrid ANN-to-SNN conversion method combining channel-wise normalization and tandem learning, obtaining an accuracy of 88.46\% on the UCF101 dataset with 200 time steps \cite{zhang2022high}. 
Recently, You et al. proposed an improved ANN-to-SNN conversion framework (scalable dual threshold mapping) to mitigate three types of conversion errors (unevenness error, clipping error and quantization error), and obtained ultra-low-latency SNNs based on the SlowFast model backbone, achieving high accuracy on UCF101 (92.94\%) and HMDB51 (67.71\%) with carefully selected hyperparameters \cite{you2024converting}. 
We observe that current SNNs for VAR have limitations for incremental training and practical application, such as complicated model construction/conversion, nontrivial input preprocessing, and long simulation time steps. 
Therefore, we directly train a novel spiking transformer for efficient and effective VAR in this paper, which is more suitable for practical deployment.

\section{Preliminary and Methodology}
In this section, we introduce the proposed SVFormer in detail. Firstly, we briefly introduce the basics of SNNs. Then, we describe the overview of SVFormer and explain the key components such as patch embedding (PE) module, local feature extractor (LFE), global self-attention (GSA) module, local pathway (LP), classification head (CH) and so on. Moreover, we introduce the method to calculate theoretical energy consumption for model inference.

\subsection{Spiking neural networks} \label{sec:snn}

The brain-inspired SNNs are regarded as the third generation of neural networks \cite{maass1997networks}, which innately exhibit temporal dynamics and communicate through binary spikes/events, attracting considerable attention and achieving significant advancements in recent years. 
SNNs offer powerful computation capability in virtue of their biological plausibility, temporal processing property with intrinsic dynamics, and energy efficiency with event-driven nature, thus considered as promising alternatives to conventional ANNs \cite{roy2019towards}. 
Specifically, the communication through sparse binary spikes allow SNNs to adopt low-power accumulate (AC) operations in convolution or linear layers, in place of power-hungry multiply-and-accumulate (MAC) operations when implemented on neuromorphic hardware, leading to high energy efficiency \cite{panda2020toward,kundu2021hire,kundu2021spike,yin2021accurate}.

\begin{figure}[h!]
\begin{center}
\includegraphics[width=\linewidth]{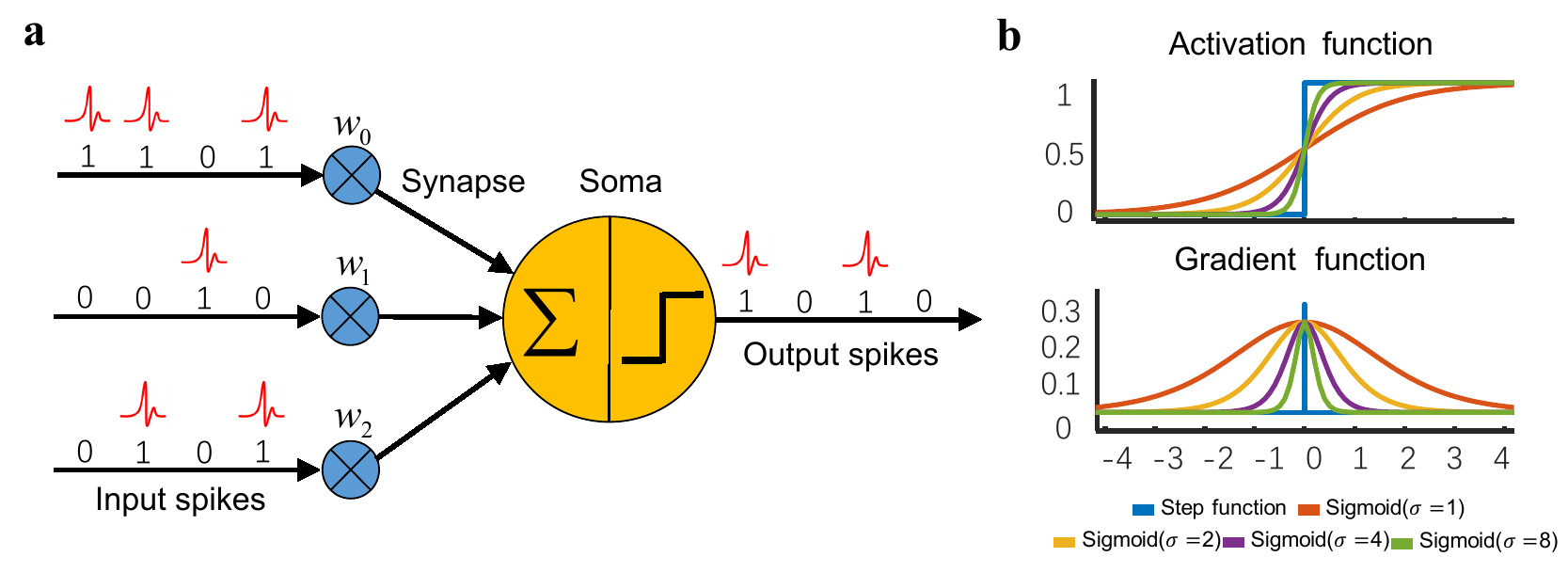}
\end{center}
\caption{(a) The scheme of a spiking neuron, of which the input and output are both binary spikes. (b) Sigmoid function approximates the Heaviside activation function of a spiking neuron, and the derivative of it can be utilized to calculate gradients during backpropagation.}\label{fig:neuron}
\end{figure}

The basic units of SNNs are spiking neurons, like ReLUs in ANNs. Leaky Integrate-and-Fire (LIF) neuron model (Fig. \ref{fig:neuron}a) is one of the most commonly adopted neuron models in SNNs \cite{fang2021deep,zhou2023spikformer,yao2023spikedriven,zhou2023spikingformer,zhang2024sglformer}. The dynamics of a LIF neuron are described as:
\begin{equation}\label{H[t]_LIF}
    H[t]=V[t-1]+\frac{1}{\tau}\left(X[t]-\left(V[t-1]-V_{reset}\right)\right),
\end{equation}
\begin{equation}\label{S[t]_LIF}
    S[t]=\Theta\left(H[t]-V_{th}\right),
\end{equation}
\begin{equation}\label{V[t]_LIF}
    V[t]=H[t]\left(1-S[t]\right)+V_{reset}S[t],
\end{equation}
where $\tau$ is the membrane time constant, $X[t]$ is the input current at time step $t$, $V_{reset}$ is the reset potential, $V_{th}$ is the firing threshold. Eq. (\ref{H[t]_LIF}) describes the update of membrane potential. Eq. (\ref{S[t]_LIF}) describes the spike generation process, where $\Theta(v)$ is the Heaviside step function: if $H[t] \geq V_{th}$ then $\Theta(v)=1$, meaning a spike is generated; otherwise $\Theta(v)=0$. $S[t]$ represents whether a neuron fires a spike at time step $t$. Eq. (\ref{V[t]_LIF}) describes the resetting process of membrane potential, where $H[t]$ and $V[t]$ represent the membrane potential before and after the evaluation of spike generation at time step $t$, respectively.
To improve the temporal representation ability of spiking neurons, trainable parameters are incorporated. For example, inspired by heterogeneous neurons in the brain, Fang et al. proposed Parametric LIF (PLIF) neuron \cite{fang2021incorporating} by using trainable membrane time constant as follows:
\begin{equation}\label{H[t]_PLIF}
    H[t]=V[t-1]+k\left(a\right)\left(X[t]-\left(V[t-1]-V_{reset}\right)\right),
\end{equation}
where $k\left(a\right)=\frac{1}{1+exp\left(-a\right)}\in\left(0,1\right)$, $\tau=\frac{1}{k(a)}$, and $a$ is the trainable parameter.

There are mainly two methods to obtain well-performing deep SNNs: ANN-to-SNN conversion and direct training through surrogate gradients.
In ANN-to-SNN conversion, a pretrained ANN is converted to an SNN by replacing the ReLU activation layers with spiking neurons and using scaling operations like weight normalization and threshold balancing \cite{cao2015spiking,hunsberger2015spiking,rueckauer2017conversion,bu2021optimal,meng2022training,wang2022signed}. To mitigate conversion error, the converted SNNs usually suffer from long simulation time steps, which causes high computational cost in practice. 
In direct training, SNNs are unfolded over the temporal dimension like RNNs and trained with backpropagation through time \cite{lee2016training,shrestha2018slayer,neftci2019surrogate}. Due to the non-differentiability of the spike generation process, the surrogate gradient method is employed for backpropagation \cite{neftci2019surrogate,lee2020enabling,fang2021deep,fang2021incorporating}. Specifically, the forward propagation utilizes Heaviside step function to generate spikes (Eq. (\ref{S[t]_LIF})), which can be approximated by differentiable functions like sigmoid and arctan functions, and the derivative of them are adopted for gradient calculation during backpropagation. Fig. \ref{fig:neuron}b illustrates the application of sigmoid function to calculate back-propagated gradients. Direct training with surrogate gradients achieves good performance with few time steps, especially on image classification tasks, thus greatly promotes the development of deep SNNs \cite{fang2021deep,fang2021incorporating,zhou2023spikformer,yao2023spikedriven,guo2023direct,yao2024spikedriven,Zhou2024DirectTH}.

\subsection{The proposed SVFormer} \label{sec:model}

\subsubsection{Overview}
\begin{figure}[h!]
\begin{center}
\includegraphics[width=\linewidth]{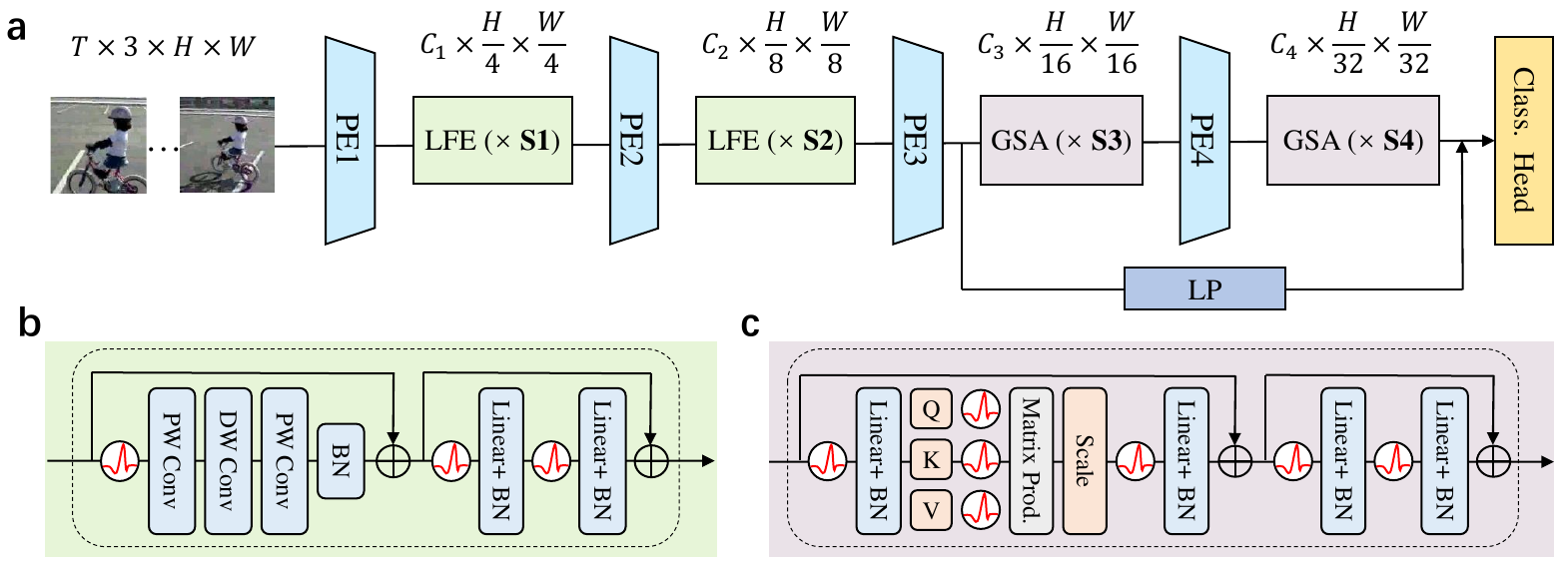}
\end{center}
\caption{(a) The overall structure of the hierarchical SVFormer, which includes four stages and one local pathway in default. (b) Structure of one local feature extractor (LFE). (c) Strcuture of one global self-attention (GSA) module.}\label{fig:svformer}
\end{figure}

The overall framework of our proposed SVFormer is shown in Fig. \ref{fig:svformer}a. Inspired from the biological brains, efficient and effective multiscale hierarchical modular structures have been widely adopted in DNNs, showing great potential for various tasks \cite{kruger2012deep,jiao2021multiscale,li2022uniformer,yao2024spikedriven}. Thus, the backbone of SVFormer is strategically structured in a hierarchical manner, comprising four stages and one local pathway, if not specified. Each of the first two stages consists of one patch embedding (PE) module and several local feature extractors (LFEs), while each of the last two stages consists of one PE module and several global self-attention (GSA) modules. The number of LFEs or GSAs in each stage is indicated as $S_i$, and the number of channels (or embedding dimension) for each stage is represented as $C_i$. If not specified, $S=[1,1,3,1]$, meaning that there are one LFE in the first two stages, three GSAs in the third stage, and one GSA in the last stage; and $C=[128, 256, 384, 512]$, meaning that the embedding dimensions for four stages are 128, 256, 384, and 512, respectively. 
Given a batch of video sequences $I\in \mathbb{R}^{B\times T\times 3\times H_{in}\times W_{in}}$, T is the length of one video sequence (usually a temporal downsampling of the original video sample), B is the batch size, and $H_{in}/W_{in}$ is the height/width of the video frame. Firstly, we reshape the input into $I\in \mathbb{R}^{T\times B\times 3\times H_{in}\times W_{in}}$, and feed one video frame into SVFormer in each time step, where T is also the total simulation time steps of SVFormer. 
After the backbone extracts spatio-temporal information from the video input, the classification head (CH) makes a decision about the action category of the input.

\subsubsection{Patch embedding module}
The PE modules play roles for feature extraction, channel dimension expansion, and patch embedding, raising channel dimension of the feature map, while downsampling the feature map in spatial dimension. The first PE module consists of a convolution (Conv) layer and a batch-normalization (BN) layer, to extract features and spatially downsample the input frame before encoding them into spikes. Other PE modules consist of a spiking neuron (SN) layer, a Conv layer, and a BN layer. For the sake of brevity, the combination of a Conv layer and a BN layer is termed as ConvBN. The computation in the last three PE modules can be expressed as $X_{PEout}=ConvBN(SN(X_{PEin}))$, and the SN operation is omitted in the first PE module as previously described.

\subsubsection{Local feature extractor}
As shown in Fig. \ref{fig:svformer}b, one LFE is composed of three cascaded Conv layers and a multi-layer perception (MLP) module, designed for extracting local spatial features. Both parts adopt residual learning with membrane shortcut \cite{yao2024spikedriven} to avoid gradient vanishing and to improve the model performance. 
To reduce the number of parameters, we adopt the style of PWConv-DWConv-PWConv like the MobileNet block \cite{howard2017mobilenets}, where PWConv (DWConv) represents pointwise (depthwise) convolution. PWConv is responsible for fusing information across the channel dimension, while DWConv with a relatively large kernel size ($5\times 5$ if not specified) for extracting spatial information in each channel. Besides, there is an SN layer ahead of the three Conv layers to transform the feature map into spikes. 
The MLP module consists of two SN-Linear-BN motifs as usual, which first expands the channel dimension by a ratio ($r=2$ if not specified) and then reduces it back to the original size, improving the model's representation capability. The combination of a Linear layer and a BN layer is termed as LinearBN. 
The computation in a LFE can be expressed as: $X = X_{LFEin} + BN(PWConv(DWConv(PWConv(SN(X_{LFEin})))))$, and $X_{LFEout} = X + MLP(X)$, where $MLP(X)=LinearBN(SN(LinearBN(SN(X))))$.

\subsubsection{Global self-attention module}
As shown in Fig. \ref{fig:svformer}c, a GSA module is composed of one spiking self-attention (SSA) module and one MLP module. Here, we adopt the SSA module which has been proven effective and efficient in \cite{zhou2023spikformer}, and use the same MLP module as that in a LFE.
The computation in an SSA module is formulated as follows: 
\begin{equation}\label{QKV}
\small
    A = SN_A(LinearBN_A(SN(X_{SSAin})), \quad A\in\{Q, K, V\}, \quad X_{SSAin}\in\mathbb{R}^{T\times B\times N\times C},
\end{equation}
\begin{equation}\label{X_{SSAout}}
\small
    X_{SSAout} = X_{SSAin} + LinearBN(SN(QK^TV*s)), \quad X_{SSAout}\in\mathbb{R}^{T\times B\times N\times C}.
\end{equation}
In an SSA module, an SN layer first transforms the input into spikes, then three parallel Linear-BN-SN motifs generate the spike-form Q, K, and V tensor, based on which the self-attention score $QK^TV*s$ is calculated through energy-efficient AC operations, where $s=1/\sqrt{64}=0.125$ is a predefined scaling factor. The self-attention score is then transformed by a Linear-BN-SN motif before being added to $X_{SSAin}$. 
In the above two equations, $N$ is the number of tokens, which is the multiplication between the height $H$ and width $W$ of the current feature map, i.e. $N=H*W$. 
Taken together, the computation in a GSA module can be expressed as: $X_{GSAout} = X_{SSAout} + MLP(X_{SSAout})$, where $X_{SSAout}$ is calculated through Eq. (\ref{QKV}) and (\ref{X_{SSAout}}).

\subsubsection{Local pathway}
To better utilize the features of different scales, we add a local pathway (LP) after the third PE module, and the output of the local pathway will be fused with the output of the last GSA module, as shown in Fig. \ref{fig:svformer}a. 
The local pathway consists of cascaded SN-DWConv-PWConv-BN layers to further extract local spatial feature in an economical way regarding parameters, and outputs a tensor with the same shape as the output of the last GSA module, which will then be concatenated along the channel dimension before sent into the classification head.
The computation of the local pathway can be expressed as $LP(X)=BN(PWConv(DWConv(SN(X)))$, where $X$ means the input of the local pathway.

\subsubsection{Classification head}
To make the most of the extracted spatio-temporal features from the backbone, we adopt a classification head (CH) similar to that in \cite{zhang2024sglformer}, which makes a learnable weighted sum of the feature map across temporal and spatial dimension, rather than simply averaging across them. 
Specifically, for the fused input $X_{CHin}\in \mathbb{R}^{T\times B\times 2C_4\times H_{out}\times W_{out}}$, we first use an SN layer to convert the feature map into spikes, and reshape it into $X_{CHinS}\in \mathbb{R}^{B\times 2C_4\times T\times H_{out}\times W_{out}}$. Then, we operate $X_{CHinS}$ with a depthwise 3D convolution layer and a BN layer, i.e. $X_{STF}=3DConvBN(X_{CHinS})$, which only slightly increases the number of parameters. Further, $X_{STF} \in \mathbb{R}^{B\times 2C_4\times 1\times 1\times 1}$ will be squeezed and operated with a linear layer to generate the classification results. Taken together, the computation in the classification head can be formulated as $Y = CH(X_{CHin}) = Linear(3DConvBN(SN(X_{CHin})))$, where the reshape and squeeze operations are omitted for brevity. 

\subsubsection{Time-dependent batch normalization}
For the sake of efficiency, a BN layer in an SNN is usually executed in a parallel way, because the calculation is not time-dependent.
The common approach is to reshape the input $X\in \mathbb{R}^{T\times B\times C\times H\times W}$ into $X\in \mathbb{R}^{TB\times C\times H\times W}$, and adopt $nn.BatchNorm2d$ to implement batch normalization \cite{zhou2023spikformer,zhou2023spikingformer,zhang2024sglformer}. Obviously, this approach utilizes the feature maps of all time steps equally and simultaneously, which is unreasonable because one cannot use future information for calculation in the current time step.
Therefore, we proposed the novel time-dependent batch normalization (TDBN) method here. Specifically, we reshape the input into $X\in \mathbb{R}^{T\times BC\times H\times W}$, and adopt $nn.BatchNorm2d$ to implement batch normalization, which treats each time step as a channel and thus independently utilizes information from different time steps.

\subsubsection{Overall computation process} 
Based on the above contents, the computation of SVFormer can be summarized as follows.
\begin{equation}
    X_{PE_iout}=PE_i(X_{PE_iin}), \quad i\in\{1,2,3,4\}
\end{equation} 
\begin{equation}
    X_{LFE_{i,j}out}=LFE_{i,j}(X_{LFE_{i,j}in}), \quad i\in\{1,2\}, \quad j\in\{1,..,S_i\}
\end{equation}
\begin{equation}
    X_{GSA_{i,j}out}=GSA_{i,j}(X_{GSA_{i,j}in}), \quad i\in\{3,4\}, \quad j\in\{1,..,S_i\}
\end{equation}
\begin{equation}
    X_{CHin}=Concate(X_{GSA_{4,S4}out}, LP(X_{PE_3out}))
\end{equation} 
\begin{equation}
    Y = CH(X_{CHin}), \quad Y\in \mathbb{R}^{B\times \#cls}
\end{equation}
In the above equations, $i$ indicates the stage index, $H$ and $W$ are the height and width of the intermediate feature maps, and $\#cls$ is the number of categories. PE, LFE, GSA, LP and CH represent the computation of patch embedding module, local feature extractor, global self-attention module, local pathway and classification head, respectively. 
The input shapes of PE, LFE, GSA, LP and CH are $X_{PE_iin}\in \mathbb{R}^{T\times B\times C_{i-1}\times H\times W}$, $X_{LFE_{i,j}in}\in \mathbb{R}^{T\times B\times C_i\times H\times W}$, $X_{GSA_{i,j}in}\in \mathbb{R}^{T\times B\times C_i\times H\times W}$, $X_{PE_3out}\in \mathbb{R}^{T\times B\times C_3\times H\times W}$, and $X_{CHin}\in \mathbb{R}^{T\times B\times 2C_4\times H_{out}\times W_{out}}$, respectively.  
The input to the first PE module is the reshaped video frames, i.e. $X_{PE_1in} = I \in \mathbb{R}^{T\times B\times 3\times H_{in}\times W_{in}}$. Besides, it should be noted that both the LFE and GSA module do not change the shape of the feature map.

\subsection{Theoretical calculation of energy consumption} \label{sec:energy}
The theoretical energy consumption of an SNN is usually calculated through multiplication between the number of MAC/AC operations and the energy consumption of each operation on predefined hardware \cite{panda2020toward,zhou2023spikformer,yao2023spikedriven,zhang2024sglformer}.
The number of synaptic operations (SOPs) are calculated as follows:
\begin{equation}\label{eq:sop}
    SOP^l=fr^{l-1} \times FLOP^l
\end{equation}
where $fr^{l-1}$ is the firing rate of spiking neuron layer $l-1$. $FLOP^l$ refers to the number of floating-point MAC operations (FLOPs) of layer $l$, and $SOP^l$ is the number of spike-based AC operations (SOPs).
Assuming the MAC and AC operations are performed on the 45nm hardware \cite{horowitz20141}, i.e. $E_{MAC}=4.6pJ$ and $E_{AC}=0.9pJ$, the energy consumption of SVFormer can be calculated as follows:
\begin{equation}\label{eq:energy}
\small
    E_{SVFormer} = E_{AC} \times \left(\sum_{i=2}^{N} SOP_{Conv/LN}^i + \sum_{j=1}^M SOP_{SSA}^j\right) + E_{MAC} \times \left(FLOP_{Conv}^1\right).
\end{equation}
$FLOP_{Conv}^1$ represents the FLOPs of the first layer before encoding input frames into spikes, $SOP_{Conv/LN}$ represents the SOPs of a convolution or linear layer, and $SOP_{SSA}$ represents the SOPs of an SSA module. $N$ is the total number of convolution layers and linear layers, and $M$ is the number of SSA modules. 
During model inference, several cascaded linear operation layers such as convolution, linear and BN layers, can be fused into one single linear operation layer \cite{zhang2024sglformer,waeijen2021convfusion}, still enjoying the AC-type operations with a spike-form input tensor.

\section{Experimental Results}
In this section, we evaluate the performance of SVFormer on two RGB video datasets (UCF101 \cite{soomro2012ucf101} and NTU-RGBD60  \cite{shahroudy2016ntu}), as well as a neuromorphic dataset (DVS128-Gesture \cite{dvs128gesture}). 
We directly train the proposed SVFormer from scratch based on the surrogate gradient method using SpikingJelly \cite{spikingjelly}, which is a popular deep learning framework for building and training SNNs. 
We compare the performance and energy consumption of SVFormer with existing SNNs or conventional ANNs to demonstrate its effectiveness and efficiency. 
Besides, we conduct ablation studies to show the effects of some network modules or simulation setups.

\subsection{Datasets and Experimental Setup}

\textbf{UCF101} includes 101 action classes with a total of 13,320 videos, which were collected from YouTube \cite{soomro2012ucf101}. These YouTube videos are recorded in unconstrained environments with cluttered background, camera motion, various illumination conditions, and beyond. These videos have a frame rate of 25 fps and a spatial resolution of $320\times 240$. The length of each video sample ranges from 1.06s to 71.04s, of which the average is 7.21s. In this study, we adopt the first official training-testing split. 

\textbf{NTU-RGBD60} contains 60 kinds of actions with a total of 56,880 samples collected with three different camera angles, which include depth, 3D skeleton, RGB and infrared sequences \cite{shahroudy2016ntu}. This study only adopts RGB videos for action recognition. Both cross-subject (C-Subject) and cross-view (C-View) splits are evaluated in this study. The C-Subject split divides the training set and testing set according to the person ID, while the C-View split divides the samples by the camera ID, as in \cite{shahroudy2016ntu}. 

\textbf{DVS128-Gesture} contains 11 different classes of gestures collected from 29 individuals under 3 different illumination conditions, which is a neuromorphic dataset collected by dynamic vision sensors \cite{dvs128gesture}. The spatial resolution of DVS128-Gesture is $128\times 128$. We first integrate the stream of events into a sequence of frames as the model input, as usually done \cite{fang2021incorporating,zhou2023spikformer,yao2023spikedriven}. 

For both RGB video datasets, the input size of the network is $224\times 224$ in both training and testing phase. Without specification, the batch size is 64, distributed across 4 Nvidia V100 GPUs. The number of training epochs is 600, with a warming-up-then-cosine-decay schedule of learning rate, of which the base value is set empirically to 0.006. 
For the DVS128-Gesture dataset, the input size of the network is $128\times 128$. We applied a 3-stage model for this small dataset. The batch size is set to 16, and the number of training epochs is 600. The learning rate is set empirically to 0.005 and decayed with a cosine schedule. 
For all three datasets, AdamW is applied as the optimizer.
Moreover, the implementation is based on the SlowFast repository \cite{feichtenhofer2019slowfast} and Uniformer repository \cite{li2022uniformer}, and common data augmentation methods like crop, flip, and random erase are applied in the training phase.

\subsection{Comparison Results}
UCF101 is a popular dataset in the VAR field, which is also commonly applied when using SNNs for VAR. Hence, we evaluate the proposed SVFormer on UCF101 and compare the performance to previous SNNs. The results are listed in Tab. \ref{tab:performance}. 
The SVFormer-base ($S=[1,1,3,1]$, $C=[128, 256, 384, 512]$) is the default network introduced in Section \ref{sec:model}, achieving a top-1 accuracy of 84.03\% on the UCF101 dataset, which is state-of-the-art among directly trained deep SNNs for VAR. And we evaluate four modifications of the base model: the shallower SVFormer-ss ($S=[1,1,2,1]$, $C=[128, 256, 384, 512]$), the thinner SVFormer-st ($S=[1,1,3,1]$, $C=[64, 128, 256, 512]$), the deeper SVFormer-dp ($S=[1,2,4,2]$, $C=[128, 256, 384, 512]$) and the wider SVFormer-wd ($S=[1,1,3,1]$, $C=[128, 256, 512, 768]$). All these modifications demonstrate lower accuracy compared to the base model. 
Three listed recurrent SNNs need nontrivial input preprocessing and 300 simulation time steps to perform the task \cite{panda2018learning,chakraborty2023heterogeneous}, which is unsuitable for practical deployment. 
The recent ANN-converted SNN, SlowFast-SDM-cv \cite{you2024converting}, only needs four simulation time steps to achieve a comparable accuracy (92.94\%) to its ANN compartment, which is critically dependent on the cautious choice of hyperparameters in the conversion process. Besides, it requires a well-trained ANN as basis and need to repeatedly process the same video clip for four times, thus not friendly for incremental training and not economic for deployment in practice. 
Further, we directly train two recently published well-performing deep SNNs, SGLFormer \cite{zhang2024sglformer} and Meta-SpikeFormer \cite{yao2024spikedriven}, on the UCF101 dataset, and the results are inferior to the SVFormer-base model. 

To validate the generalizability of the proposed SVFormer, we evaluate SVFormer on a large RGB dataset (NTU-RGBD60) and a neuromorphic dataset (DVS128-Gesture). 
To the best of our knowledge, SVFormer is the first SNN model that has ever been assessed on the NTU-RGBD60 dataset. Hence, we compare it to two recently published well-performing ANNs tested on RGB frames \cite{siddiqui2024dvanet,reilly2023just}. The results in Tab. \ref{tab:performance} show that SVFormer's accuracy is slightly lower, which is acceptable for the substantial savings of energy consumption. 
For DVS128-Gesture, We applied the SVFormer-3stg model with one local stage and two global stages, where $S=[1,2,1]$ and $C=[64,128,256]$. At the same time, we exclude the local pathway for it. The accuracy of the SVFormer-3stg model is 97.92\%, which is comparable to mainstream SNNs but with significantly fewer parameters (Tab. \ref{tab:performance}).

\begin{table*}[!htbp]
  \centering
  \caption{Comparison results of top-1 accuracy on UCF101, DVS128-Gesture and NTU-RGBD60. $\dag$ indicates our implementations. The best-performing results are highlighted in bold. SVFormer-base is SOTA of directly trained deep SNNs on UCF101.} 
    \begin{tabular}{p{2.5cm}<{\raggedright}p{4.0cm}<{\raggedright}p{1.0cm}<{\centering}p{1.0cm}<{\centering}p{2.8cm}<{\centering}}
    \toprule
    \multicolumn{1}{l}{Dataset} &{Model} & Param\newline{}(M)  & {Time Steps} & {Top-1 Acc (\%)} \\
    
    \midrule
    \multirow{12}{*}{UCF101}  
    & RSNN-reservoir-DA \cite{panda2018learning} &40.40   &300 & 81.30 \\
    & RSNN-HeNHeS-STDP \cite{chakraborty2023heterogeneous} &-   &300 & 77.53 \\
    & RSNN-HeNB-BP \cite{chakraborty2023heterogeneous} &-   &300 & 84.32 \\
    \cmidrule(lr){2-5}
    & RSNN2s-tandem-cvt \cite{zhang2022high} &-   &200 & 88.46 \\
    & SlowFast-SDM-cvt \cite{you2024converting} &- &4 & \textbf{92.94}\\
    \cmidrule(lr){2-5}
    & SGLFormer-8-384 \cite{zhang2024sglformer} $\dag$ &11.76 &16  & 74.70\\
    & Meta-SpikeFormer \cite{yao2024spikedriven} $\dag$ &13.81 &16   & 83.66\\
    \cmidrule(lr){2-5}
    & SVFormer-base &13.80 &16 & \textbf{84.03} \\
    & SVFormer-ss &12.53   &16 & 83.61 \\
    & SVFormer-st &8.77   &16 & 80.15 \\
    & SVFormer-dp &17.72   &16 & 80.25 \\
    & SVFormer-wd &24.07   &16 & 81.23 \\

    \midrule
    \multirow{8}{*}{DVS128-Gesture}
    & ConvNet-PLIF \cite{fang2021incorporating} &-   &20 & 97.57 \\
    & RSNN-HeNHeS-STDP \cite{chakraborty2023heterogeneous} &-   &100 & 96.54 \\
    & RSNN-HeNB-BP \cite{chakraborty2023heterogeneous} &-   &100 & 98.12 \\
    & Spikformer-2-256 \cite{zhou2023spikformer} &2.57   &16 & 98.30 \\
    & Spikingformer-2-256 \cite{zhou2023spikingformer} &2.57   &16 & 98.30 \\
    & SD-Transformer-2-256 \cite{yao2023spikedriven} &2.57   &16 & \textbf{99.30} \\
    & SGLFormer-3-256 \cite{zhang2024sglformer} &2.17   &16 & 98.60 \\
    \cmidrule(lr){2-5}
    & SVFormer-3stg & 1.88   &16 & 97.92 \\

    \midrule
    \multirow{3}{*}{NTU-RGBD60}  
    & DVANet (ANN) \cite{siddiqui2024dvanet} &-   &- & 93.40(CS)/\textbf{98.20}(CV) \\
    & $\pi$-ViT (ANN) \cite{reilly2023just} &-   &- & \textbf{94.00(CS)}/97.90(CV) \\
    \cmidrule(lr){2-5}
    & SVFormer-base & 13.76   &16 & 88.12(CS)/94.68(CV) \\
    
    \bottomrule
    \end{tabular}%
  \label{tab:performance}%
\end{table*}%

\begin{table*}[!htbp]
  \centering
  \caption{Performance of SVFormer-base on the UCF101 dateset under different noise conditions.}
    \begin{tabular}{p{2.0cm}<{\raggedright}p{1.0cm}<{\centering}p{1.0cm}<{\centering}p{1.0cm}<{\centering}p{1.0cm}<{\centering}p{1.0cm}<{\centering}p{1.0cm}<{\centering}p{1.0cm}<{\centering}}
    \toprule
    Noise condition & Null & \multicolumn{3}{c}{Gaussian noise (a)} & \multicolumn{3}{c}{Salt-and-pepper noise (P)} \\
      & - & 0.1 & 0.5 & 1 & 0.1 & 0.2 & 0.3 \\
    \midrule
    Top1-acc (\%) & 84.03 & 82.26 & 77.13 & 64.76 & 75.36 & 66.32 & 55.17 \\
    \bottomrule
    \end{tabular}%
  \label{tab:robustness}%
\end{table*}%

\begin{figure}[h!]
\begin{center}
\includegraphics[width=\linewidth]{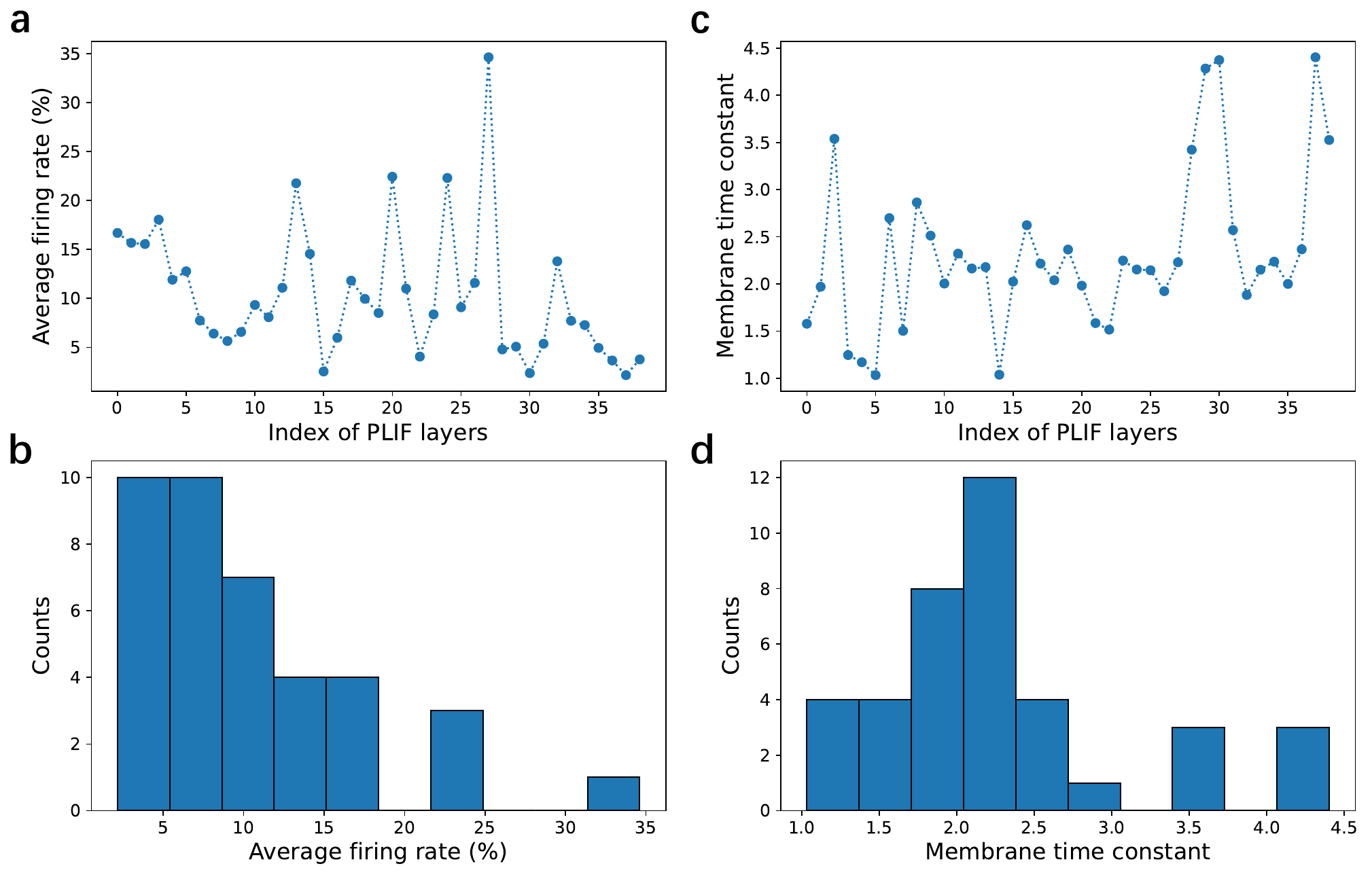}
\end{center}
\caption{Average firing rates (a, b) during model inference on the UCF101 dataset, and learned membrane time constants (c, d), for all PLIF layers, showing the trend of both variables as the network goes deeper (a, c), and the corresponding histograms (b, d).}\label{fig:fr_tau}
\end{figure}

As described in Section \ref{sec:snn}, a main strength of SNNs is energy efficiency, which can be attributed to the sparsity of spikes and the spike-driven communication in SNNs. In this section, we calculate the average theoretical energy consumption of the SVFormer-base model in inference for one video clip from UCF101, with the method described in Section \ref{sec:energy}. 
Firstly, we calculate the number of MAC operations (FLOPs) of convolution and linear layers; then, we count the average firing rate of each spiking neuron layer, and convert corresponding MAC operations (FLOPs) to AC operations (SOPs) by Eq. (\ref{eq:sop}); finally, we calculate the theoretical energy consumption by Eq. (\ref{eq:energy}). 
The average firing rate of each spiking neuron layer is shown in Fig. \ref{fig:fr_tau} (a, b), which indicates the sparsity of spikes during model inference. 
The FLOPs of the SVFormer-base model's ANN counterpart is 229.163G if executed for 16 repetitions (as 16 time steps in the SNN version), of which the energy consumption is 1054.148mJ, where all the spiking neurons are replaced by ReLUs. 
According to the above method, the remaining FLOPs of the SVFormer-base model is 0.700G, and the SOPs is 20.760G, thus the energy consumption is 21.904mJ. 
Obviously, the energy cost of SVFormer-base is much lower than its ANN counterpart, validating the energy efficiency of SNNs. 
Further, the ratio of the energy cost of SVFormer-base to that of its ANN counterpart is 1.99\% (21:1054), which is much lower than that of SlowFast-SDM-cv (98:128=76.56\%) \cite{you2024converting}, showing advantages of directly trained SNNs compared to ANN-converted ones. 

\begin{figure}[h!]
\begin{center}
\includegraphics[width=\linewidth]{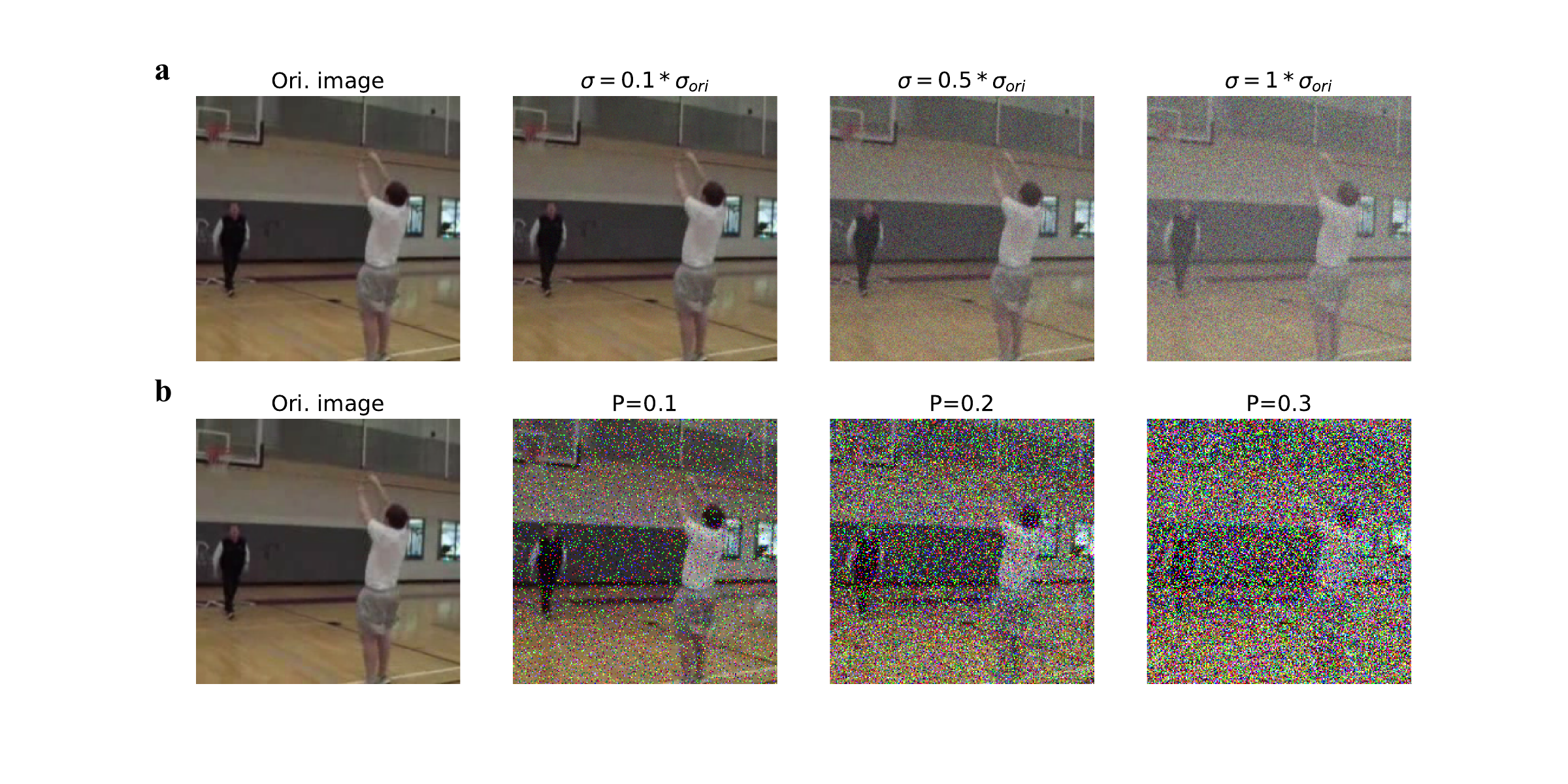}
\end{center}
\caption{(a) An exemplar frame with different level of Gaussian noise ($N(0, \sigma)$). (b) An exemplar frame with different level of salt-and-pepper noise, where $P$ means the sprinkling probability.}\label{fig:robustness}
\end{figure}

Moreover, to test the robustness of the model, we evaluate the trained SVFormer-base model with noisy frames from the UCF101 dataset. Here, we applied two commonly adopted noise types for images, i.e. Gaussian noise and salt-and-pepper noise. For the Gaussian type, we add noise with zero mean and different level of standard deviation $\sigma = a*\sigma_{ori}$ to the original frame, where $\sigma_{ori}$ indicates the \textit{std} of the original frame. For the salt-and-pepper noise, we randomly transform a pixel into the highest or lowest value of the current frame with a predefined probability $P$. Fig. \ref{fig:robustness} demonstrates the effects of different level of noise with an exemplar frame. 
The results in Tab. \ref{tab:robustness} demonstrate that SVFormer exhibits resilience to a moderate level of noise, sustaining commendable performance. However, when confronted with excessive noise that markedly degrades frame quality, there is a substantial decline in the model's performance.

\subsection{Ablation studies}
In this subsection, we conduct some ablation studies based on the UCF101 dataset to show the effects of some network modules or simulation setups. The results are listed in Tab. \ref{tab:ablation}, where batch size are adjusted according to the model size. 
First, classification accuracy of a SVFormer without the local pathway is 83.29\%, which is slightly lower than the original 84.03\%, showing that the local pathway improves the model's representational power for VAR. 
Second, classification accuracy of a SVFormer with LIF neurons is only 80.89\%, demonstrating that learnable membrane time constants of spiking neurons (PLIF) enhance the model's temporal processing capability. As shown in Fig. \ref{fig:fr_tau} (c, d), the learned membrane time constants are variable, helping to better represent temporal information, which is also consistent with heterogeneous neurons observed in biological brains. Besides, the membrane time constants of the deeper layers are on average larger than those of the shallower layers, implying that deeper layers integrate spatio-temporal information for longer duration. 
Third, without using time-dependent batch normalization to utilize information from different time steps independently, the model' classification accuracy decrease from 84.03\% to 80.65\%, indicating the effectiveness of the proposed time-dependent BN. 
Fourth, we apply different number of frames (i.e. T=8 or T=24) as network input to test the effects of input video length, where T is the number of frames sampled from the original video sample. And the simulation duration of the SNN model is adjusted to 8 or 24 steps accordingly. Both 8-frame and 24-frame inputs exhibit lower accuracies compared to the original 16-frame input, indicating the necessity of finding a suitable input length to balance accuracy and computational cost.

\begin{table*}[!htbp]
  \centering
  \caption{Ablation studies on the UCF101 dateset.}
    \begin{tabular}{p{5.0cm}<{\raggedright}p{1.0cm}<{\centering}p{1.0cm}<{\centering}p{1.5cm}<{\centering}}
    \toprule
    \multicolumn{1}{l}{Architecture} & Param\newline (M)  & {Batch size} & {Top-1 Acc (\%)} \\
    \midrule
    \multicolumn{1}{l}{\multirow{1}{*}{SVFormer Base (T=16)}}  & 13.80 &64 & 84.03 \\
    \midrule
    \multicolumn{1}{l}{\multirow{1}{*}{Base - Local pathway}}  & 12.32 &64 & 83.29 \\
    \multicolumn{1}{l}{\multirow{1}{*}{Base + (PLIF $\rightarrow$ LIF)}}  & 13.80 &64 & 80.89 \\
    \multicolumn{1}{l}{\multirow{1}{*}{Base + (BN $\rightarrow$ TDBN)}}  & 13.35 &64 & 80.65 \\
    \midrule
    \multicolumn{1}{l}{\multirow{1}{*}{Base + (T=8)}}  & 12.76 &64 & 81.26 \\
    \multicolumn{1}{l}{\multirow{1}{*}{Base + (T=24)}}  & 14.84 &40 & 81.81 \\
    \midrule
    \multicolumn{1}{l}{\multirow{1}{*}{Base + 3D clip input (16/4)}}  & 12.69 &56 & 74.20 \\
    \multicolumn{1}{l}{\multirow{1}{*}{Base + 3D clip input (20/4)}}  & 12.89 &40 & 74.09 \\
    \bottomrule
    \end{tabular}%
  \label{tab:ablation}%
\end{table*}%

Additionally, when utilizing SNNs to process time sequences such as videos, addressing the alignment of temporal resolution becomes a critical concern that warrants attention. 
In this work, we mainly adopt the frame-by-frame approach, i.e. the input length (number of frames) is equal to the simulation duration (number of time steps) of the SNN model, which processes one frame per time step. 
Here, we also test the clip-by-clip approach, where the SNN model processes a clip of frames in each time step. Firstly, we need to modify the 2D-Conv and 2D-BN layers in the model into 3D ones accordingly, to process a 3D video clip. Then, we separate a sampled video sequence with 16 or 20 frames into 4 parts uniformly, meaning that the model runs 4 time steps and processes one clip with 4 or 5 frames in each time step. Finally, we train the model from scratch and test its performance. The results in Tab. \ref{tab:ablation} demonstrate that the frame-by-frame approach is superior to the clip-by-clip approach for the UCF101 dataset.

\section{Conclusion}
In this paper, we propose the directly trained SVFormer, which integrates local feature extraction, global self-attention, and the intrinsic dynamics, sparsity, and spike-driven nature of SNNs, effectively and efficiently learning spatio-temporal representation for VAR. 
We evaluate SVFormer on two RGB datasets (UCF101, NTU-RGBD60) and a neuromorphic dataset (DVS128-Gesture). The experimental results demonstrate that SVFormer achieves comparable performance to mainstream models for VAR tasks in a more efficient way. Specifically, SVFormer achieves state-of-the-art (top-1 accuracy of 84.03\%) among directly trained deep SNNs on UCF101, with ultra-low power consumption (21 mJ/video). 
These results verify SVFormer’s strong representation capability, showing that the multiscale spatio-temporal feature-extraction characteristic endows it with great potential as a backbone for diverse video tasks when equipped with properly designed task heads. 
However, there are certain limitations associated with this pioneering endeavor. On the one hand, the recognition accuracy lags behind the state-of-the-art of traditional ANNs. One possible solution is to try large-scale self-supervised pretraining for SNNs, which has already shown great success for ANNs \cite{tong2022videomae,wang2023videomae,wei2022masked}. 
On the other hand, although a video clip is processed frame-by-frame in SVFormer, the length of the clip is predefined for the ease of network implementation, meaning that the model needs to process all frames before making a decision. One can try to achieve speed-accuracy tradeoff in the model as a biological brain \cite{bogacz2022speed}, i.e. the model may stop at any step if a decision is made based on predefined criteria, which is more flexible and efficient. 

\begin{credits}
\subsubsection{\ackname} %
This study was supported by grants 62206141 and 62236009 from the National Natural Science Foundation of China, and grant PCL2021A13 from Peng Cheng Laboratory.
\subsubsection{\discintname}
The authors have no competing interests to declare that are relevant to the content of this article.
\end{credits}
%
%
%
\bibliographystyle{splncs04}
\bibliography{snn4video}

\end{document}